# Automation of Pedestrian Tracking in a Crowded Situation

**S. Saadat and K. Teknomo**

Ateneo De Manila University, Philippines
Corresponding author: samansad@gmail.com

**Abstract** Studies on microscopic pedestrian requires large amounts of trajectory data from real-world pedestrian crowds. Such data collection, if done manually, needs tremendous effort and is very time consuming. Though many studies have asserted the possibility of automating this task using video cameras, we found that only a few have demonstrated good performance in very crowded situations or from a top-angled view scene. This paper deals with tracking pedestrian crowd under heavy occlusions from an angular scene.

Our automated tracking system consists of two modules that perform sequentially. The first module detects moving objects as blobs. The second module is a tracking system. We employ probability distribution from the detection of each pedestrian and use Bayesian update to track the next position.

The result of such tracking is a database of pedestrian trajectories over time and space. With certain prior information, we showed that the system can track a large number of people under occlusion and clutter scene.

## Introduction

Measurement of individual pedestrian trajectories, which is part of microscopic pedestrian field of study, is useful in a variety of situations. In business environments, information about customer's movement derives better knowledge on shopping behavior and preferences. In crowded locations (such as sports venues, public transport station, religious events), pedestrians' movement influences safety concerns.

One major approach to obtain pedestrian trajectories is to use video camera and track each individual pedestrian from the video camera from the time he/she is detected on the scene until he/she leaves the scene. However, given the normal video frame rate of 25-30 images per second, to track hundreds of individual pedestrians manually would requires herculean effort despite its accuracy. Better approach is to automate the tracking system.

While such automation in tracking in video camera is not new in the literatures of computer vision (e.g. see [1], [2], and [3] for good survey on the field), most of them are performed only for a relatively few pedestrians. Relatively few recent papers (e.g. [4]), however, reported to track pedestrian on a crowded situation in short-medium range sometime up to maximum distance of 8 meters from the camera. Most of the person detection technique works better in a well controlled environment such as laboratory setting with a small amount of people [2] whereas other techniques require more than one camera with overlapping fields of view in order to cope with the pedestrian segmentation problem in the a dense crowd [5].

In the real environments, while walking, the people form groups and then separate from one another, and more over they have shadows [1] and the object, especially, people undergo a change in their shapes while moving and their motion is not constant that cause the tracking is a difficult problem [3].

This paper describes our attempt to automate such tracking. Tracking pedestrians in a crowded situation is a hard problem because pedestrian does not move as a rigid body. Human body goes through a large range of variation during walking. In crowded situation, due to the angle of the camera, the bodies of pedestrians which are farther from the camera are occluded by other nearer pedestrians. Furthermore, in outdoor scene, the lighting condition is uncontrolled and it may create shadow that hinders correct detection of pedestrians.

Our goal is to track all pedestrian's trajectories in the scene and save the data of tracking as a NTYX table where, N is the pedestrian number, T is time in video, and X and Y are the coordinate position of the pedestrian in the image that readily converted into scene coordinate.

Our automated tracking system consists of two modules that perform sequentially. The first module is called pedestrian detection. The second module is pedestrian tracking. Therefore, this paper is organized as follow. First, we explain the two modular part of our system: pedestrian detection and pedestrian tracking. Then, we discuss the result of our system compared to the ground truth data. Finally we conclude the paper.

**Pedestrian Detection**

The first module detects moving objects as blobs. For this module we have two options:
1. *Without prior information*: this option creates generic background image and then subtracts it from the video data to obtain the foreground scene.
2. *With prior information*: as we as the pedestrians to wear red hat, we only detect the red hat based on color.

We employ background modeling to create background image when the pedestrian detection system run without prior information. The background modeling is also dealing with the difficulty to detect the shadow and part of the background. We assume the background model follows Gaussian distribution, thus we compute the average and variance of the image sequence using recursive time average

**Equation 1**

$$\mu_t = \frac{t-1}{t}\mu_{t-1} + \frac{I_t}{t}$$

$$\sigma_t^2 = \alpha\sigma_{t-1}^2 + (1-\alpha)(I_t - \mu_t)^2$$

The background image is updated every time using the following formula

**Equation 2**

$$B_t = \mu_t + \eta\sigma_t$$

The foreground image is obtained through background subtraction and thresholded to get the binary image.

**Equation 3**

$$F_t = (I_t - B_t) > \phi$$

The second option does not employ any background modeling because we have prior information that all the pedestrians we track wear the same red color of hats. In this case, our goal is to extract only the red hat information by removing other unnecessary color information. Subtracting green channel from the red channel will produce color of high reddish value and very low greenish value at the same time. That is equivalent to the range of violet to red. Since we also have prior knowledge that the road color was dark, the Blue channel does not carry much information to be subtracted from the red.

Therefore our object detection algorithm is simply a subtraction of green channel from the red channel and put some threshold to make them binary image.

**Equation 4**

$$F_t = (R_t - G_t) > \phi$$

Higher threshold value produces cleaner blob but it will also remove small blobs. After a sequence of morphological image operations (dilate, open and close) to clean the noise, we get the blobs ready for pedestrian tracking.

## Pedestrian Tracking

The tracking procedure begins when the moving objects on the scene have been detected as blobs. These binary blobs are used as masks to the color image to obtain the color blobs. Utilizing these blobs, we can find the basic features such as color histogram, area, and center of gravity. Our tracking algorithm is based on the Bayesian update and the probabilities are computed for each blob on each frame. In each set of computation, we consider three consecutive frames (i.e. frame t-2, t-1 and current frame t).

All of these features of the blobs are stored into a multidimensional feature matrix for each two consecutive frames. Each column in the matrix indicates a blob in the lower frame number (i.e. frame t-2 in matrix of t-2 and t-1, or frame t-1 in matrix t-1 and t), and each row represents a blob in the higher frame number (i.e. frame t in matrix t-1 and t). The other dimensions of the matrix are used to store each features.

Then, these features are thresholded to get binary multidimensional matrix. The entries of the binary matrix represent possibilities that a blob in one frame will become the equivalent blob in the next frame because they represent the same pedestrian. Equivalent blobs can either exactly the same blob, or merge with other blobs, or separated into several blobs. It is the task of tracking system to detect these possibilities through the computation of probabilities.

The probability of each blob is computed based on derived features from the feature matrix. We have four derived features namely entropy difference, movement angle, speed and distance. The derived features are explained as follow.

One unique feature of our algorithm is the usage of entropy of each blob. The entropy is computed based on the color histogram of the blob. To calculate the entropy we use the following equation.

**Equation 5**

$$E_c = -\sum \frac{H_c}{A_c} * \log(\frac{H_c}{A_c})$$

Where, H is the color histogram of the blob when there is non-zero value (complete black) and subscript c is the index of three color channels and A is the area of the blob in color channel. We have three values in the entropy of each blob representing each color channel and finally we sum the entropy of the three channels to get entropy of the blob. Once entropy of the blob is computed, we can obtain the entropy difference by subtracting the entropy of the blob on the higher frame number from the entropy of the lower frame number, and then take the absolute value.

**Equation 6**

$$\Delta E = |E_h - E_l|$$

The next derived features that will be used to compute probability value are movement angle, speed and distance. Given the center coordinate of the blob X, we can compute the directional angle and speed (distance between two blobs in three consecutive frames) as suggested by [6].

**Equation 7**

$$\theta = w_1(1 - \frac{\overline{X_{i,t-2}X_{j,t-1}}\cdot\overline{X_{j,t-1}X_{k,t}}}{\|\overline{X_{i,t-2}X_{j,t-1}}\|\|\overline{X_{j,t-1}X_{k,t}}\|})$$

**Equation 8**

$$v = w_2(1 - 2\frac{\sqrt{\|\overline{X_{i,t-2}X_{j,t-1}}\|\|\overline{X_{j,t-1}X_{k,t}}\|}}{\|\overline{X_{i,t-2}X_{j,t-1}}\| + \|\overline{X_{j,t-1}X_{k,t}}\|})$$

The subscripts i, j, k indicate index of the blobs at time t-2, t-1 and t consecutively and w1 and w2 are weight parameters. The next derived feature is distance between blobs between two blobs in two consecutive frames.

The values of the derived features indicate the dissimilarity of two blobs, means if the values are zero then they are more similar and higher values means less similarity. Therefore we need to normalize the derived features to represent similarity. The normalization of the derived features is using Equation 9 to make slow decrease for small value of the derived feature and then rapidly decrease if the value of the derive features is large.

**Equation 9**

$$f_{l,h,m} = \sqrt{1 - (f_{l,h,m} / \varphi_m)^2}$$

Having all the derived features that stored into multidimensional matrix, we can flatten the multidimensional matrix into 2-dimensional matrix by setting a linear combination with the weight of each feature. Then we also put threshold for each feature to obtain a binary matrix.

**Equation 10**

$$\sum_m \omega_{l,h,m} f_{l,h,m} < \varphi$$

For each two consecutive frame, we have one binary matrix that we called possibility matrix. Based on the possibility matrix, we can derive a tree structure of possibilities that a blob in one frame is equivalent to another blob in other frames.

Figure 1 illustrates a hypothetical probability tree with all the necessary naming convention of the probabilities. The probability of the second and third level of the probability tree is computed as follow

**Equation 11**

$$P_{i|j} = \frac{\sum_m \omega_{i,j,m} f_{i,j,m}}{\sum_i \sum_j \sum_m \omega_{i,j,m} f_{i,j,m}}$$

**Equation 12**

$$P_{k|i,j} = \frac{\sum_m \omega_{j,k,m} f_{j,k,m}}{\sum_j \sum_k \sum_m \omega_{j,k,m} f_{j,k,m}}$$

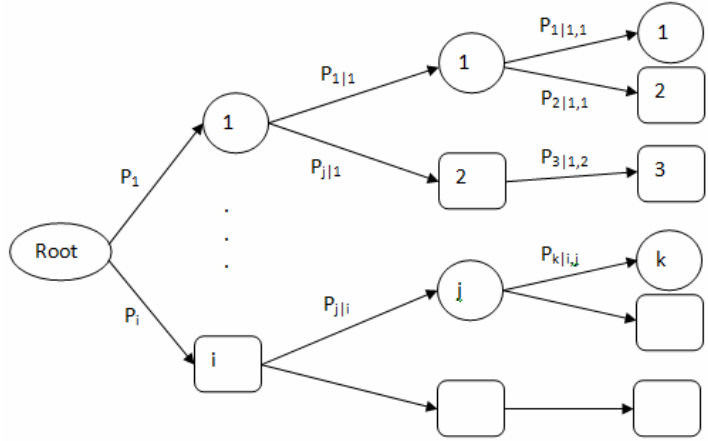

**Fig. 1 Probability tree name convention**

After calculating all the probabilities, we calculate the posterior probability using Bayes Theorem for each leaf in tree as follow

**Equation 13**

$$P_{i|j,k} = \frac{P_{j|i} P_{k|i,j} P_i}{\sum_i \sum_j \sum_k P_{j|i} P_{k|i,j} P_i}$$

The prior probability is computed based on the update of features on the second level of the probability tree

**Equation 14**

$$P_i \leftarrow P_j = \sum_i P_{j|i}$$

First we find the maximum posterior probability in possibility matrix. Once we found match blob, then we make the row and the column of the matched blob into to zero. The procedure is repeated until the possibility matrix becomes a zero matrix.

## Results & Analysis

The result of our detection system without prior information is relatively good to detect pedestrian in not so crowded situation as illustrated in Figure 3 with only 80% accuracy. The top left image is the original image of the video scene. The blobs are shown on the top right and in the bottom left shows the detected pedestrians. The left bottom show the trajectories tracking results in the image coordinates.

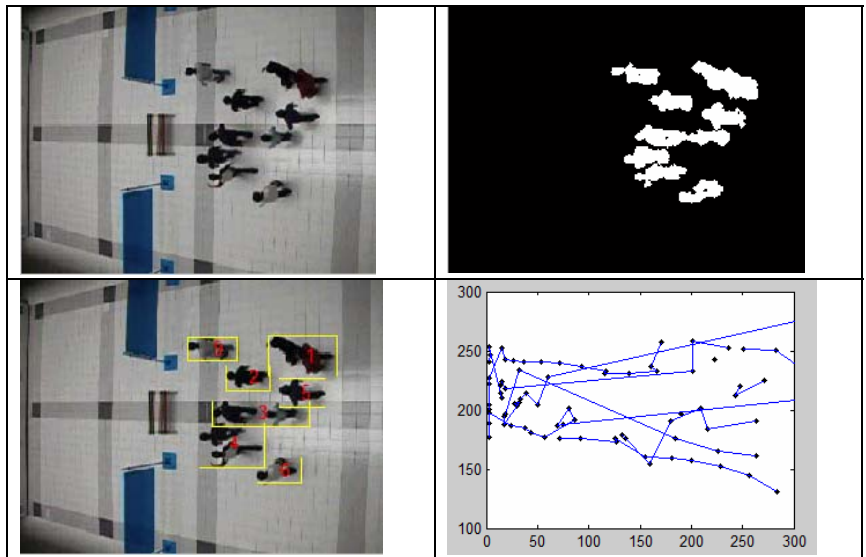

**Fig. 2 Example of tracking without prior information**

For crowded scene, the tracking without prior information produces low level of accuracy below 40%.

The crowded scene as illustrated in Figure 3 was taken during pedestrian experiment in Ateneo de Manila University, Philippines for the study of microscopic pedestrian. The video was taken from the third floor of a building (about 10 meter height) at unknown angle. The camera was not calibrated.

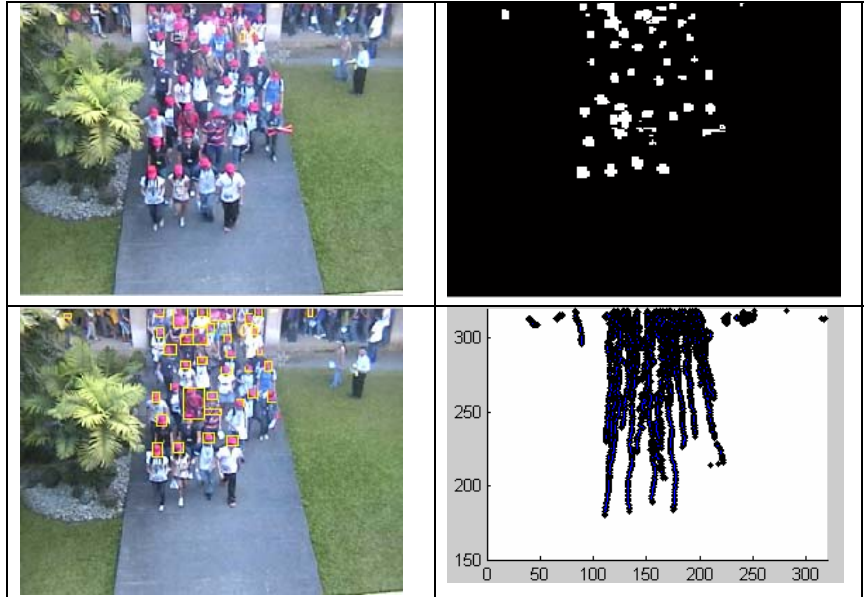

**Fig. 3 Example of tracking with prior information**

We use prior information to detect the blobs, our algorithm was able to track with much higher result and in some way more than 90% and the algorithm was able to find lost pedestrian even after 10 frames of occlusion.

This algorithm can track pedestrian very well because the detection part was almost perfect while the Bayesian update make the algorithm to be get better result over the longer time of tracking. Each frame that a pedestrian track it will increase its posterior probability and it eventually produces a better result. At the same time, however, our algorithm needs a very large amount of memory to save all blobs images and the computational speed becomes very slow.

## Conclusion

Tracking pedestrian in a crowded situation is a hard problem due to occlusion and clutters. We attempted to solve such problem by employing uncalibrated video camera from a top view angle. We compare two options of detections: one without prior information and the other using prior information about the scene. Then, we use the same tracking algorithm to track them. It is showed that prior information significantly increases the accuracy of detection.

We also presented a new algorithm to calculate the probability of tracking system using Bayesian Update in a probability tree. It has been shown that the tracking system can detect more than 90% of the pedestrian even in crowded situation and even when they move zigzag.

Improvement of such system for both part detection and tracking to have faster and more reliable system would be subject to further research study.

## Acknowledgement

This project supported by Pedestrian Research Group of Ateneo De Manila University